\documentclass{article}
\usepackage{deeprobot_paper,times}

\usepackage{amsmath,amsfonts,bm}

\def\eqref#1{equation~\ref{#1}}

\def\1{\bm{1}}

\DeclareMathAlphabet{\mathsfit}{\encodingdefault}{\sfdefault}{m}{sl}
\SetMathAlphabet{\mathsfit}{bold}{\encodingdefault}{\sfdefault}{bx}{n}

\usepackage{amssymb,booktabs,multirow,array,enumitem,xurl}
\usepackage{hyperref}
\newcommand{\cmark}{\ensuremath{\checkmark}}
\newcommand{\pmark}{\ensuremath{\triangle}}
\newcommand{\benchname}{LynxScenes}
\newcommand{\paperauthor}[2]{\mbox{#1\textsuperscript{#2}}}
\newcommand{\authorsep}{\penalty0\hskip0.75em\relax}

\title{\benchname{}: A Large-Scale Real-World Benchmark for Outdoor Wheel-Legged Robot Perception, Navigation, Imagination and Reconstruction}
\author{%
  \paperauthor{Jun Tian}{1,2}\authorsep
  \paperauthor{Hao Wu}{3}\authorsep
  \paperauthor{Ruo Chen}{4}\authorsep
  \paperauthor{Yiyi Liao}{4}\authorsep
  \paperauthor{Hongyi Lin}{5,6}\authorsep
  \paperauthor{Jincheng Yu}{5}\authorsep
  \paperauthor{Chao Li}{1,*}\authorsep
  \paperauthor{Yingjie Niu}{1,*,$\dagger$}%
}
\affiliation{%
  \small
  \textsuperscript{1}\,Deep Robotics\quad
  \textsuperscript{2}\,University of Illinois at Urbana-Champaign\\
  \textsuperscript{3}\,The Hong Kong University of Science and Technology\\
  \textsuperscript{4}\,Zhejiang University\quad
  \textsuperscript{5}\,Tsinghua University\\
  \textsuperscript{6}\,Massachusetts Institute of Technology\\[0.25em]
  {\footnotesize\textsuperscript{*}\,Corresponding authors\quad
  \textsuperscript{$\dagger$}\,Project Leader}
}
\paperdate{}
\projecturl{https://developer.deeprobotics.cn/research/articles/lynx-scenes/index.html}
\hypersetup{
  pdftitle={LynxScenes: A Large-Scale Real-World Benchmark for Outdoor Wheel-Legged Robot Perception, Navigation, Imagination and Reconstruction},
  pdfauthor={Jun Tian, Hao Wu, Ruo Chen, Yiyi Liao, Hongyi Lin, Jincheng Yu, Chao Li, Yingjie Niu},
  pdfsubject={Real-world wheel-legged robot dataset and benchmark},
  pdfkeywords={LynxScenes, wheel-legged robot, navigation, BEV perception, world model, scene reconstruction}
}

\begin{document}
\makebrandheader
\begin{deeprobotfrontmatter}
  \maketitle
  \begin{abstract}
Wheel-legged robots operating in outdoor spaces shared with humans require accurate perception, socially compliant navigation, future-scene reasoning, and scene reconstruction.
However, existing datasets typically support these capabilities in isolation and rarely provide dense spatial and behavior-level supervision on real wheel-legged platforms.
We introduce \benchname{}, a large-scale real-world dataset and benchmark collected with a LYNX M20 wheel-legged robot.
It contains approximately 100K 30-s navigation clips spanning 833.6 h and 3,294.7 km, with 3.0M annotated frames.
The synchronized recordings include four fisheye cameras, dual LiDARs, IMU, odometry, and robot states.
Annotations comprise 20.82M oriented 3D boxes across 27 categories, 13.89M road-structure annotations, and 20 navigation instructions organized into three layers and aligned with continuous future trajectories.
\benchname{} supports four complementary tasks: instruction-conditioned trajectory prediction, BEV-based perception of 3D objects and road structures, future scene prediction, and multi-view scene reconstruction.
For trajectory prediction, we further introduce a log-replay-based pseudo-closed-loop evaluator that measures safety, progress, and instruction compliance beyond trajectory imitation.
Extensive evaluations across representative navigation, perception, world-model, and reconstruction methods reveal substantial remaining challenges in close-range outdoor robot navigation.
The dataset and evaluation toolkit are planned for public release.
  \end{abstract}
\end{deeprobotfrontmatter}

\section{Introduction}
\label{sec:introduction}

\begin{figure}[htbp]
\centering
    \includegraphics[width=0.8\columnwidth]{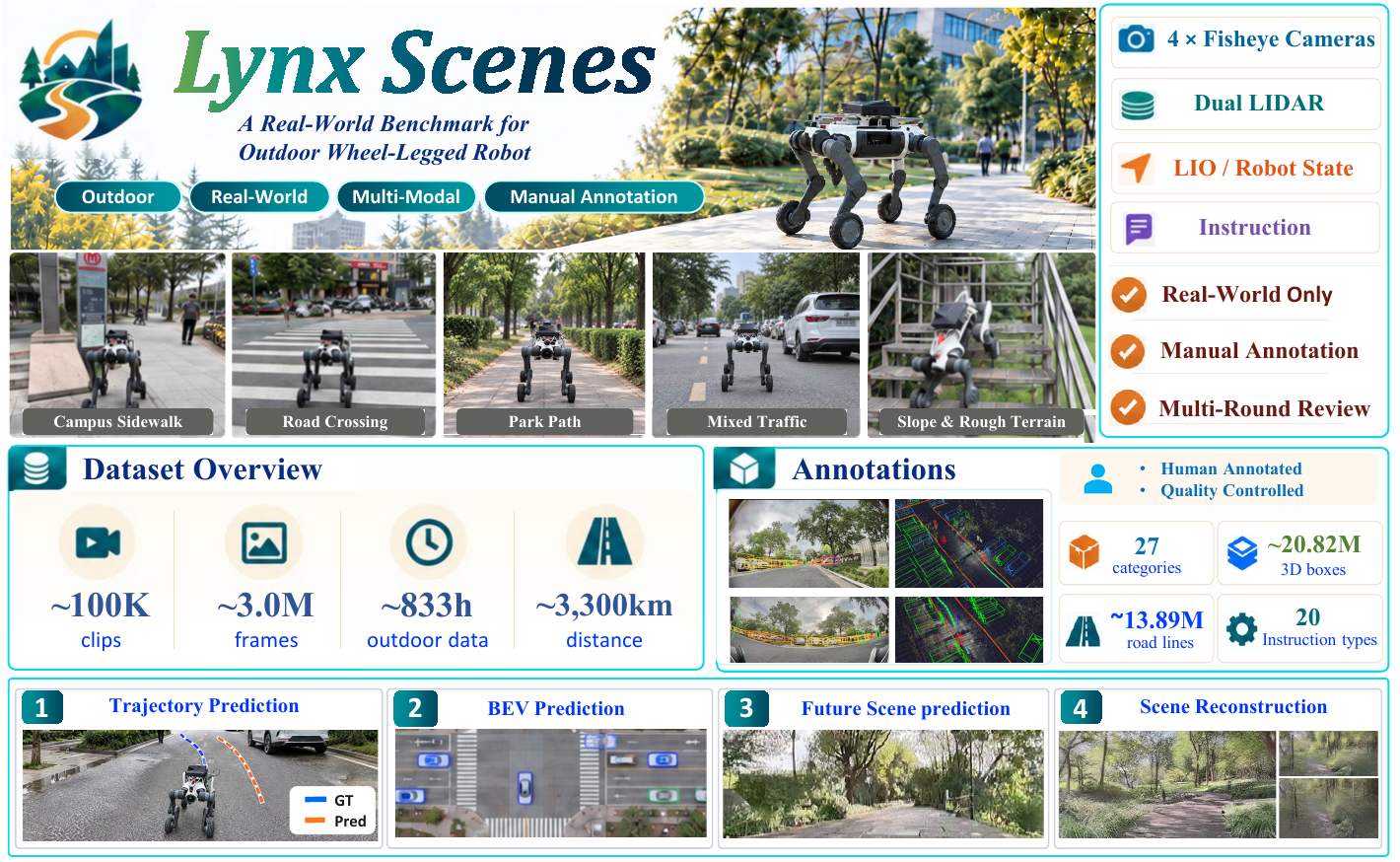}
    \caption{
    \textbf{Overview of \benchname{}.}
    The benchmark integrates large-scale real-world outdoor data collection,
    multimodal sensing, dense spatial, behavioral annotations, and four
    downstream tasks: instruction-conditioned trajectory prediction, BEV perception,
future scene prediction, and multi-view scene reconstruction.
    }
    \label{fig:overview}
\end{figure}

Wheel-legged robots are increasingly deployed in open outdoor environments shared with humans, including campuses, intersections, and parks.
Combining efficient wheeled locomotion with the ability to traverse uneven terrain, they are promising platforms for delivery, inspection, and assistive applications.
Unlike road vehicles operating within lane-structured environments or indoor robots navigating regular layouts, these robots move through loosely structured spaces from low, continuously changing viewpoints and often interact closely with pedestrians, cyclists, and other dynamic agents.
Reliable navigation therefore requires not only safe motion but also socially compliant behavior in shared public spaces.

Such operation requires several tightly coupled capabilities: perceiving nearby agents and environmental structures, anticipating future scene evolution, generating continuous trajectories under platform-specific motion constraints, and reconstructing representative scenes for replay and analysis.
Yet these capabilities are typically studied using separate datasets and evaluation protocols.
Evaluating them on shared real-world observations enables consistent analysis under the same embodiment, environment, and interaction distribution.
In autonomous driving, datasets such as KITTI-360~\citep{kitti360}, nuScenes~\citep{nuscenes}, Waymo~\citep{waymo}, and
Argoverse~2~\citep{argoverse2} have enabled multiple autonomy problems to be studied on common real-world data.
However, their vehicle-mounted viewpoints, lane-centric topology, and interaction regimes differ substantially from those of wheel-legged robots operating in pedestrian spaces.

Existing robot-navigation and language-conditioned navigation datasets provide only subsets of the required sensing and supervision, as summarized in Table~\ref{tab:dataset_comparison}.
VLN benchmarks provide instruction-following supervision ~\citep{r2r,rxr,vlnce,touchdown}, but their simulated or graph-based interfaces largely abstract away physical robot dynamics and lack dense spatial annotations.
Robot-centric datasets reduce this embodiment gap but remain fragmented across perception, social navigation, traversability, localization, and mapping~\citep{jrdb,scand,recon,gnd,m3ed,diter}.
Among closely related efforts, GrandTour~\citep{grandtour} provides rich multimodal measurements for legged-robot perception and state estimation but lacks behavior-level language instructions, whereas CoT-VLNBench~\citep{cotvlnbench} provides instructions, future trajectories, and 3D object boxes but lacks metric annotations of navigation-relevant road structures.
The field therefore still lacks a large-scale real-world wheel-legged robot dataset that jointly provides multimodal observations, dense spatial annotations, behavior-level instructions, and continuous future trajectories.

To address this gap, we introduce \benchname{} (Fig.~\ref{fig:overview}), a large-scale real-world dataset and benchmark for socially compliant outdoor wheel-legged robot navigation.
Data are collected with a DEEP Robotics LYNX M20 operated by trained human operators following a predefined social-compliance protocol.
\benchname{} contains approximately 100K non-overlapping 30-s navigation clips spanning 833.6 h and 3,294.7 km, with 3.0M annotated frames.
It provides synchronized multimodal sensing and robot states, together with 20.82M oriented 3D boxes across 27 categories, 13.89M road-structure annotations, 20 behavior-level navigation instructions in three layers,
and aligned continuous future trajectories.

Using the same sequences and training/evaluation splits, we benchmark four complementary tasks: instruction-conditioned trajectory prediction, BEV-based perception of 3D objects and road structures, future scene prediction, and multi-view scene reconstruction for visual replay and as a foundation for future simulation-based testing.
For trajectory prediction, we further introduce a log-replay-based
pseudo-closed-loop evaluator inspired by NAVSIM~\citep{navsim}.
Its Quadruped Navigation Score (QNS) evaluates collision avoidance,
walkable-area compliance, clearance, progress, and instruction adherence,
complementing conventional trajectory-displacement errors.
We evaluate 10 VLN/VLA policies, 4 BEV perception models, 6 world models, and 5 reconstruction methods.
The results reveal substantial room for improvement in close-range interactions and complex outdoor environments.
The dataset, standardized splits, and evaluation toolkit are planned for public release.

Our main contributions are summarized as follows: \textbf{(1)} We introduce \benchname{}, a large-scale real-world multimodal dataset collected on a wheel-legged robot under a standardized social-compliance protocol.
It jointly provides socially compliant expert demonstrations, dense 3D object and road-structure annotations, structured behavior-level instructions, and continuous future trajectories.
To the best of our knowledge, it is the largest publicly reported real-world wheel-legged robot navigation dataset with expert trajectories and dense 3D scene annotations. \textbf{(2)} We establish a unified benchmark spanning instruction-conditioned trajectory prediction, BEV perception, future scene prediction, and multi-view scene reconstruction, together with a log-replay-based pseudo-closed-loop evaluator that measures safety, navigation quality, and instruction adherence. \textbf{(3)} We systematically evaluate representative methods across all four tasks, establishing reproducible baselines, identifying persistent challenges in close-range outdoor navigation, and providing standardized data splits and a public evaluation toolkit.

\begin{table}[!tbp]
    \centering
    \caption{
    Comparison with representative robot-navigation datasets
    and benchmarks.
    ``Temp. ID'' denotes persistent object identities across temporal
    observations;
    ``Road/Trav. GT'' denotes explicit road-structure, map, or
    traversability supervision;
    and ``Cmd.$\rightarrow$Traj.'' denotes direct conditioning of
    continuous trajectory or waypoint prediction on semantic navigation
    instructions.
    Scale is reported using the primary unit provided by each dataset.
    ``N/R'' indicates that a physical duration or traveled distance is
    not reported, while ``--'' denotes not applicable or unavailable.
    }
    \label{tab:dataset_comparison}

    \renewcommand{\arraystretch}{1.25}
    \setlength{\tabcolsep}{1.2pt}
    \fontsize{7.2}{8.8}\selectfont

    \begin{tabular*}{\linewidth}{@{\extracolsep{\fill}}lcccccccccccc@{}}
        \toprule

        \textbf{Dataset}
        & \textbf{Real}
        & \textbf{Outdoor}
        & \textbf{Legged}
        & \textbf{LiDAR}
        & \deeptablehead{3D}{Obj.}
        & \deeptablehead{Temp.}{ID}
        & \deeptablehead{Road/}{Trav.}
        & \deeptablehead{Sem.}{Cmd.}
        & \deeptablehead{Cmd.$\rightarrow$}{Traj.}
        & \textbf{Scale}
        & \textbf{Duration}
        & \textbf{Distance}
        \\

        \midrule
        \multicolumn{13}{l}{
        \textit{Real-world multimodal and robot-navigation datasets}
        }\\
        \midrule

        M2DGR~\citep{m2dgr}
        & \cmark
        & \cmark
        & --
        & \cmark
        & --
        & --
        & --
        & --
        & --
        & 36 sequences
        & 3.8\,h
        & 10.7\,km
        \\

        RECON~\citep{recon}
        & \cmark
        & \cmark
        & --
        & \cmark
        & --
        & --
        & --
        & --
        & --
        & $>$5K trajectories
        & $>$40\,h
        & N/R
        \\

        SCAND~\citep{scand}
        & \cmark
        & \cmark
        & \cmark
        & \cmark
        & --
        & --
        & --
        & --
        & --
        & 138 trajectories
        & 8.7\,h
        & 40.2\,km
        \\

        GND~\citep{gnd}
        & \cmark
        & \cmark
        & --
        & \cmark
        & --
        & --
        & \cmark
        & --
        & --
        & 10 campuses
        & 11.1\,h
        & 53\,km
        \\

        M3ED~\citep{m3ed}
        & \cmark
        & \cmark
        & \cmark
        & \cmark
        & --
        & --
        & --
        & --
        & --
        & 57 sequences
        & 3.4\,h
        & N/R
        \\

        DiTer~\citep{diter}
        & \cmark
        & \cmark
        & \cmark
        & \cmark
        & --
        & --
        & \cmark
        & --
        & --
        & 4 sequences
        & 1.0\,h
        & 1.65\,km
        \\

        GrandTour~\citep{grandtour}
        & \cmark
        & \cmark
        & \cmark
        & \cmark
        & --
        & --
        & --
        & --
        & --
        & 49 sequences
        & $>$5\,h
        & $>$10\,km
        \\

        \midrule
        \multicolumn{13}{l}{
        \textit{Language-conditioned navigation benchmarks}
        }\\
        \midrule

        RxR~\citep{rxr}
        & --
        & --
        & --
        & --
        & --
        & --
        & --
        & \cmark
        & --
        & 126K pairs
        & --
        & --
        \\

        VLN-CE~\citep{vlnce}
        & --
        & --
        & --
        & --
        & --
        & --
        & --
        & \cmark
        & --
        & 16.8K episodes
        & --
        & --
        \\

        Touchdown~\citep{touchdown}
        & --
        & \cmark
        & --
        & --
        & --
        & --
        & --
        & \cmark
        & --
        & 9.3K examples
        & --
        & --
        \\

        VLN-PE~\citep{vlnpe}
        & \pmark
        & --
        & \cmark
        & --
        & --
        & --
        & --
        & \cmark
        & \pmark
        & Physical benchmark
        & --
        & --
        \\

        CoT-VLNBench~\citep{cotvlnbench}
        & \cmark
        & \cmark
        & \cmark
        & \cmark
        & \cmark
        & \cmark
        & --
        & \cmark
        & \cmark
        & 60 sequences / 175K frames
        & 5\,h
        & N/R
        \\

        \midrule

        \textbf{Ours}
        & \cmark
        & \cmark
        & \cmark
        & \cmark
        & \cmark
        & \cmark
        & \cmark
        & \cmark
        & \cmark
        & \textbf{100K clips / 3.0M frames}
        & \textbf{833.6\,h}
        & \textbf{3,294.7\,km}
        \\

        \bottomrule
    \end{tabular*}

    \vspace{1mm}

    \footnotesize
    \pmark~denotes partial or related support that is not directly
    equivalent to the corresponding task formulation.
\end{table}

\section{Related Work}
\label{sec:related_work}

\subsection{Real-World Robot Navigation Datasets}
\label{subsec:robot_navigation_datasets}

Real-world robot datasets provide complementary supervision for
localization, perception, and navigation.
M2DGR~\citep{m2dgr} and M3ED~\citep{m3ed} emphasize multimodal sensing
and state estimation, while RECON~\citep{recon},
SCAND~\citep{scand}, GND~\citep{gnd}, DiTer~\citep{diter}, and
GrandTour~\citep{grandtour} provide various forms of trajectory,
demonstration, traversability, or robot-state supervision.
However, existing datasets typically expose only subsets of the
spatial and behavioral information required to jointly study
perception and navigation.
In contrast, \benchname{} combines synchronized multimodal observations,
persistent 3D objects, explicit road structures, structured navigation
instructions, and continuous future trajectories within the same
real-world wheel-legged robot dataset.

\subsection{Instruction-Conditioned Navigation and Planning}
\label{subsec:command_navigation}

Classical VLN benchmarks study instruction following primarily in
simulated or image-based environments.
R2R~\citep{r2r}, RxR~\citep{rxr}, and VLN-CE~\citep{vlnce} focus on
indoor navigation, whereas Touchdown~\citep{touchdown} considers
outdoor street-view imagery.
Recent benchmarks move closer to physical navigation:
VLN-PE~\citep{vlnpe} incorporates embodiment and realistic robot
dynamics, while CoT-VLNBench~\citep{cotvlnbench} provides real
quadruped navigation data with language instructions, 3D object
annotations, and continuous trajectories. However, CoT-VLNBench does not provide explicit road-structure geometry for trajectory-level spatial-safety evaluation.
Our benchmark further aligns continuous future trajectories with a
three-layer instruction taxonomy together with explicit object and road-structure
annotations, enabling instruction-conditioned trajectory prediction and
trajectory-level spatial-safety evaluation.
Inspired by replay-based planning evaluation such as
NAVSIM~\citep{navsim}, we additionally evaluate predicted trajectories
using safety, progress, and instruction-compliance criteria rather than
trajectory similarity alone.

\subsection{BEV Perception and Spatial Scene Understanding}
\label{subsec:bev_related}

Autonomous-driving datasets such as
nuScenes~\citep{nuscenes}, Waymo~\citep{waymo}, and
Argoverse~2~\citep{argoverse2} have established large-scale benchmarks
for 3D and BEV perception.
Methods including BEVFormer~\citep{li2022bevformer},
BEVFusion~\citep{liu2022bevfusion}, and
MapTRv2~\citep{MapTRv2} further model surrounding objects and road
geometry in BEV.
These benchmarks are primarily designed around vehicle-scale traffic
and driving-oriented map structures, whereas wheel-legged robots
operate close to pedestrians, curbs, sidewalks, and crosswalks.
\benchname{} therefore jointly evaluates nearby 3D objects and
navigation-relevant road structures in a robot-centric BEV frame,
providing spatial supervision that is directly aligned with the local
constraints governing safe robot motion.

\subsection{World Models and Neural Scene Reconstruction}
\label{subsec:world_reconstruction_related}

Generative world models such as GAIA-1~\citep{gaia1},
Navigation World Models~\citep{nwm}, and Cosmos~\citep{cosmos,cosmos3}
predict future visual observations from current scenes and motion or
action conditions.
In parallel, 3D Gaussian Splatting~\citep{3dgs} and recent
feed-forward reconstruction methods enable efficient scene
reconstruction and novel-view synthesis.
Most existing evaluation is conducted in driving or generic visual
settings rather than low-viewpoint outdoor robot navigation.
\benchname{} evaluates both future scene prediction and
multi-view scene reconstruction on the same real-world robot
data, providing complementary benchmarks for temporal scene evolution
and viewpoint-dependent scene replay. This complementary design supports both predictive reasoning about future observations and scene-level analysis for outdoor robot navigation.

Detailed discussion of the related methods is provided in
Appendix~\ref{sec:additional_related_work}.

\begin{figure}[htbp]
    \centering
    \includegraphics[width=0.9\columnwidth]{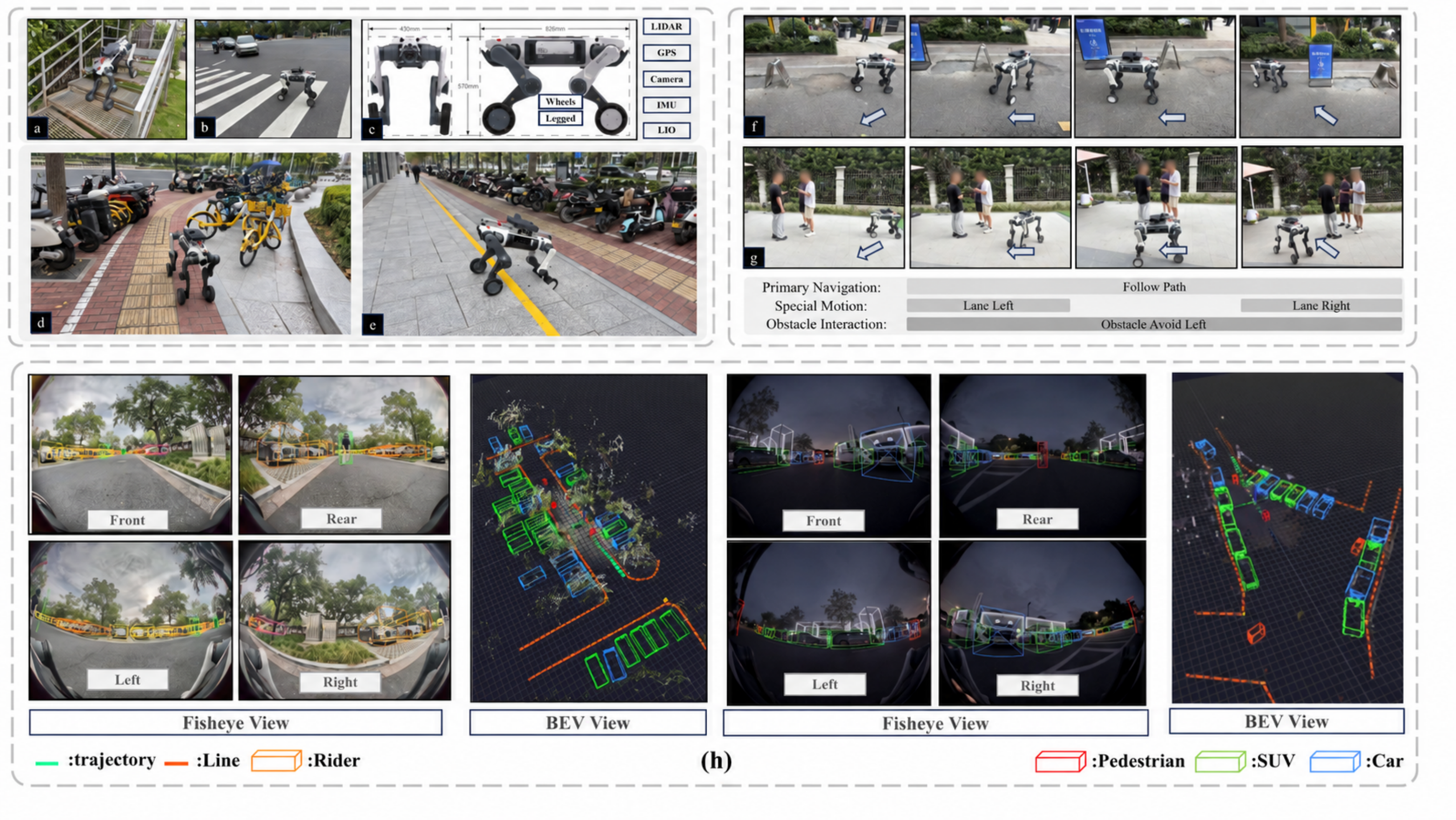}
\caption{
Overview of the data-collection platform, representative navigation behaviors,
and multimodal annotations in our benchmark.
(a)--(e) show the DEEP Robotics LYNX M20 wheel-legged quadruped, its physical
configuration, onboard sensing setup, and representative outdoor navigation
scenarios.
(f)--(g) illustrate robot behaviors together with temporally aligned semantic
instructions covering primary navigation, special motion, and obstacle interaction.
(h) visualizes synchronized four-view fisheye observations at different times and the corresponding
bird's-eye-view annotations, including surrounding 3D objects, road structures,
and the ego trajectory.
}
    \label{fig:robot_setup}
\end{figure}

\section{Dataset}
\label{sec:dataset}

\benchname{} is collected with a physical wheel-legged robot and
organized into approximately 100K non-overlapping navigation clips of
30\,s each, amounting to 3.0M annotated frames, 833.6\,h of navigation,
and 3,294.7\,km of traveled distance.
Each clip contains synchronized surround-view images, LiDAR point
clouds, and robot states, while each annotated frame provides
oriented 3D object boxes, road-structure annotations, structured
navigation instructions, and a continuous future trajectory.
The same recordings support instruction-conditioned trajectory prediction,
BEV-based perception of 3D objects and road structures,
future scene prediction, and multi-view scene reconstruction,
as defined in Sec.~\ref{sec:benchmark}.

\subsection{Platform and Sensor Setup}
\label{subsec:platform}

The data-collection platform is a DEEPRobotics LYNX M20 wheel-legged
quadruped (Fig.~\ref{fig:robot_setup}).
It carries four fisheye cameras facing front, rear, left, and right,
each recording $1920\times1536$ images at 10\,Hz, and two 96-channel
RoboSense Airy LiDARs mounted at the front and rear.
Together, the LiDARs provide $360^{\circ}$ horizontal and
$90^{\circ}$ vertical coverage to a range of 60\,m at 10\,Hz.

The platform additionally records IMU measurements, 6-DoF
LiDAR-inertial odometry (LIO), and robot states.
The robot-state stream provides the longitudinal velocity $v_x$,
lateral velocity $v_y$, and yaw rate $\omega_z$, and is used to
construct the trajectory supervision described in
Sec.~\ref{subsec:trajectory}.
All geometric quantities are represented in the robot-centric
Front--Left--Up (FLU) coordinate system, with $x$ forward,
$y$ left, and $z$ up.
The complete sensor specifications are provided in
Appendix~\ref{app:sensor_configuration}.

\subsection{Data Collection and Synchronization}
\label{subsec:sensor_setup}

All data are collected through human teleoperation of the physical robot
in real outdoor environments, including sidewalks, streets, intersections,
campuses, parks, narrow passages, slopes, open grounds, and rural roads.
The recordings cover both structured and unstructured terrain, static
obstacles, and frequent interactions with pedestrians, riders, and vehicles.

Annotated frames are sampled at 1~Hz from the synchronized multimodal
recordings.
Camera--LiDAR calibration, temporal alignment, and stream synchronization
details are provided in Appendix~\ref{app:sensor_configuration}.
Data collection is performed by trained operators following predefined
socially compliant navigation guidelines; the collection protocol and
quality-control procedure are detailed in
Appendix~\ref{app:data_quality_control}.

\subsection{Trajectory Construction}
\label{subsec:trajectory}

Reference future trajectories are constructed from the robot-state
stream.
Given the body-frame longitudinal velocity $v_x$, lateral velocity
$v_y$, and yaw rate $\omega_z$, the robot motion is temporally
integrated and transformed into the robot coordinate frame at the
current anchor timestamp.
The resulting trajectory is resampled to 10\,Hz, providing a sequence
of future planar poses associated with each annotated frame.
The complete integration formulation is given in
Appendix~\ref{app:trajectory_construction}.

\subsection{Spatial and Navigation Annotations}
\label{subsec:annotations}

Beyond trajectory supervision, the benchmark provides rich spatial and semantic annotations to support multiple key tasks in wheel-legged robot navigation. These annotations capture surrounding objects, road structures, and navigation intent, enabling both perception- and planning-oriented evaluation. The benchmark provides 27 object categories, four road-structure
classes, and 20 navigation instructions organized into three layers.
The complete annotation taxonomy is provided in
Appendix~\ref{app:annotation_taxonomy}.

\textbf{3D objects.}
Every object around the robot is
annotated with an oriented 3D bounding box, one of 27 semantic
categories, a persistent
instance identity, a dynamic state, and a velocity.
Identities are maintained across frames and temporally
consecutive clips, which supports long-term association and
per-object collision accounting.

In total the dataset contains 20.82M boxes, 6.94 per annotated frame
on average.

\textbf{Road structure.}
We annotate four road-structure classes, represented as polylines or closed crosswalk contours.

These annotations encode the accessible area of the robot and are
the basis of the Walkable-Area Compliance in
Sec.~\ref{sec:benchmark}.
The dataset contains 13.89M road-structure annotations, 4.63 per frame
on average.

\textbf{Navigation instructions.}
Each frame is tagged with structured instructions from three independent
layers.
\emph{Primary navigation} describes the dominant route
intent and covers the entire timeline, \emph{Special motion} marks robot-specific maneuvers and
\emph{Obstacle interaction} marks how the robot responds to
an obstacle.

Instructions within a layer are mutually exclusive, while instructions from
different layers may overlap in time.
Unlike joystick or velocity commands, these labels describe
behavior-level intent, and their alignment with the future trajectory
is what makes instruction-conditioned prediction and behavior-wise
evaluation possible.

\textbf{Annotation procedure and quality control.}
All annotations and recorded expert demonstrations undergo manual quality
control for annotation completeness, consistency and compliance
with the predefined socially compliant navigation guidelines.
Detailed review and filtering procedures are provided in
Appendix~\ref{app:data_quality_control}.

\subsection{Dataset Statistics and Splits}
\label{subsec:dataset_statistics}

\begin{figure}[t]
    \centering
    \includegraphics[width=0.85\textwidth]{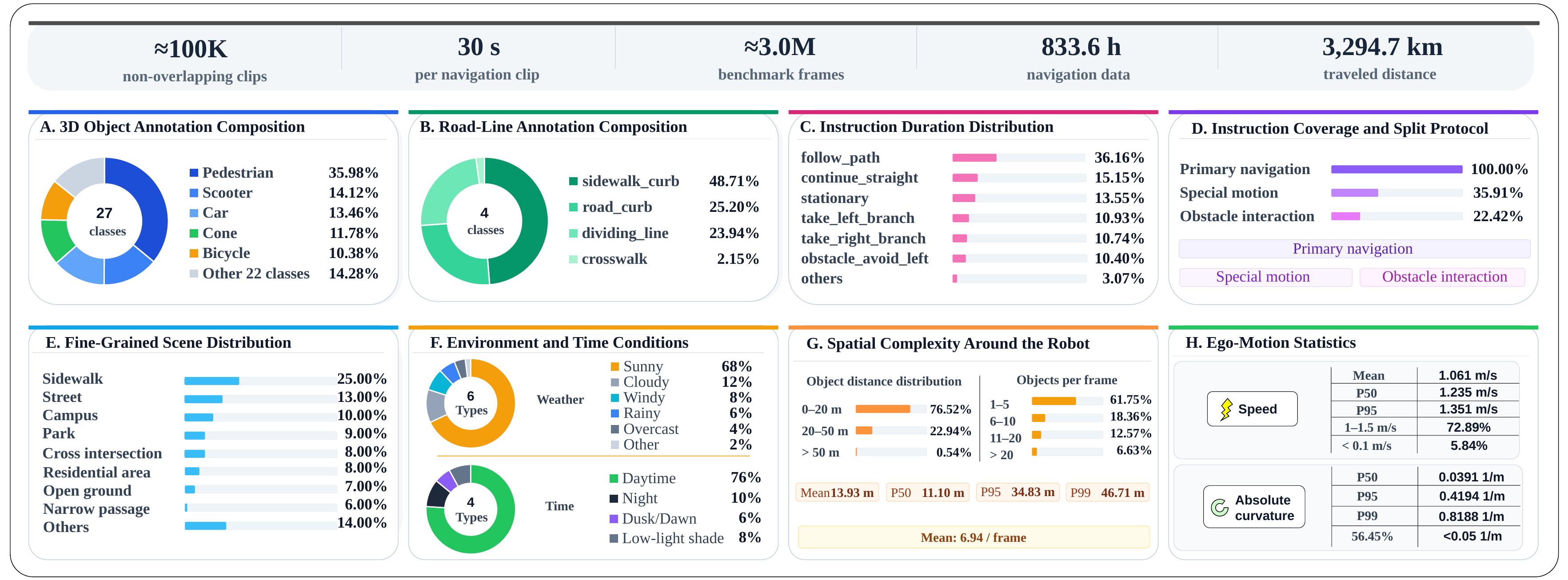}
    \caption{
    \textbf{Dataset statistics.}
    (A) Composition of 3D object annotations.
    (B) Composition of road-structure annotations.
    (C) Distribution of navigation instructions.
    (D) Coverage of the three instruction layers.
    (E) Scene-type distribution.
    (F) Environment and time-of-day conditions.
    (G) Object distance and per-frame object counts.
    (H) Ego speed and trajectory curvature.
    }
    \label{fig:dataset_statistics}
\end{figure}

As summarized in Fig.~\ref{fig:dataset_statistics}, \benchname{} covers a wide
range of real-world outdoor navigation conditions, including diverse scene
types, environmental and illumination conditions, road structures, and
human--robot interaction scenarios. Beyond environmental diversity, the
dataset spans substantial variation in object categories, spatial complexity,
semantic navigation behaviors, and ego-motion patterns, ranging from routine
path following to special-motion and obstacle-interaction instructions. Such broad
coverage supports systematic evaluation across heterogeneous navigation
conditions and interaction complexities. The 30-s clip is used as the atomic
split unit, with clips divided 8:1:1 into training, validation, and test sets
while approximately preserving the navigation-instruction distribution.
To avoid temporal and scene leakage, clips originating from the same
continuous collection session are assigned to the same split.

\section{Benchmark Tasks and Metrics}
\label{sec:benchmark}
To validate the capability of our benchmark to support key wheel-legged robot navigation problems, we establish four representative benchmark tasks. These tasks span navigation prediction, spatial perception,
future scene modeling, and scene reconstruction in real-world outdoor
environments.
\subsection{Instruction-Conditioned Trajectory Prediction}

We formulate the primary benchmark task as
instruction-conditioned continuous trajectory prediction. For each anchor $t$, a model receives four historical timestamps
$\{-3,-2,-1,0\}$ s, each with synchronized front, rear, left, and right fisheye
images (16 images total), the current motion state $\mathbf{s}_t=[v_x,v_y,\omega_z]$,
and the three-layer semantic instruction. It predicts the complete future trajectory
$\hat{\mathbf{T}}_t\in\mathbb{R}^{10\times3}$ over 1 s at 10 Hz, where each pose
is $[x_{\mathrm{forward}},y_{\mathrm{left}},\psi]$ in the anchor Body-FLU frame.

We report imitation accuracy using ADE, FDE, AHE, FHE, and complement it with
a pseudo-closed-loop replay inspired by NAVSIM~\citep{navsim}. Annotated agents
are replayed against the predicted ego trajectory and six interpretable sub-scores
measure No Collision (NC), Walkable-Area Compliance (WAC), Time-to-Collision
(TTC), Clearance (CL), Ego Progress (EP), and Instruction Compliance (IC). They are
combined as
\begin{equation}
    \mathrm{QNS}=\mathrm{NC}\times\mathrm{WAC}\times
    \frac{w_{\mathrm{EP}}\mathrm{EP}+w_{\mathrm{TTC}}\mathrm{TTC}+w_{\mathrm{CL}}\mathrm{CL}+w_{\mathrm{IC}}\mathrm{IC}}
    {w_{\mathrm{EP}}+w_{\mathrm{TTC}}+w_{\mathrm{CL}}+w_{\mathrm{IC}}}.
    \label{eq:qns}
\end{equation}
The replay construction, footprint geometry, interpolation rules, thresholds,
and instruction-specific compliance criteria are detailed in Appendix~\ref{sec:appendix_navigation}.

\subsection{BEV-Based Perception of 3D Objects and Road Structures}
\label{subsec:bev_tasks}

To evaluate spatial understanding for wheel-legged robot navigation,
we consider BEV-based perception of surrounding 3D objects and
navigation-relevant road structures.

\textbf{For 3D object detection}, models predict confidence-scored oriented boxes
in the current robot frame and are evaluated on Car and Pedestrian using
center-distance-based mAP and HDS~\citep{nuscenes}. \textbf{For road-structure detection}, Road Curb, Sidewalk Curb, and
Road--Sidewalk Dividing Line form an ordinary-line class, while Crosswalk
is represented as a closed-contour class and evaluated using
Chamfer-distance-based AP. Detailed matching rules, thresholds, and metric definitions are provided
in Appendix~\ref{sec:appendix_bev}.

\subsection{Future Scene Prediction}

\label{subsec:world_model_task}

To evaluate future scene prediction, we assess whether models can generate
the future egocentric visual evolution of outdoor robot navigation scenes.
Given an initial rectified front-view image $\mathbf{I}_0$, models generate
a future RGB video and, when supported, additionally receive future motion
conditions; text prompts are used when required by the native model
interface.

We evaluate complementary aspects of generation quality, including
perceptual quality, temporal quality, appearance consistency, and
distributional realism, using Q-Align-I/V~\citep{qalign},
CLIP-IQA+~\citep{clipiqa}, LAION-Aes~\citep{laion_aes},
DOVER-T~\citep{dover}, VBench temporal metrics~\citep{vbench},
CLIP-I~\citep{openclip}, and FVD~\citep{fvd}.
Model details, temporal normalization, and metric definitions
are provided in Appendix~\ref{sec:appendix_world_model}.

\subsection{Multi-View Scene Reconstruction}
\label{sec:gs_task}

To evaluate multi-view scene reconstruction, models reconstruct the observed
environment from multiple context views and render RGB images at held-out
viewpoints.

We benchmark representative Gaussian-splatting reconstruction methods and
evaluate novel-view rendering using complementary pixel-, structural-, and
perceptual-quality metrics.
Detailed context-view settings, evaluation protocols, and metric definitions
are provided in Appendix~\ref{sec:appendix_gs}.

\section{Experiments}
\label{sec:experiments}
\begin{figure*}[t]
    \centering
    \includegraphics[width=0.85\textwidth]{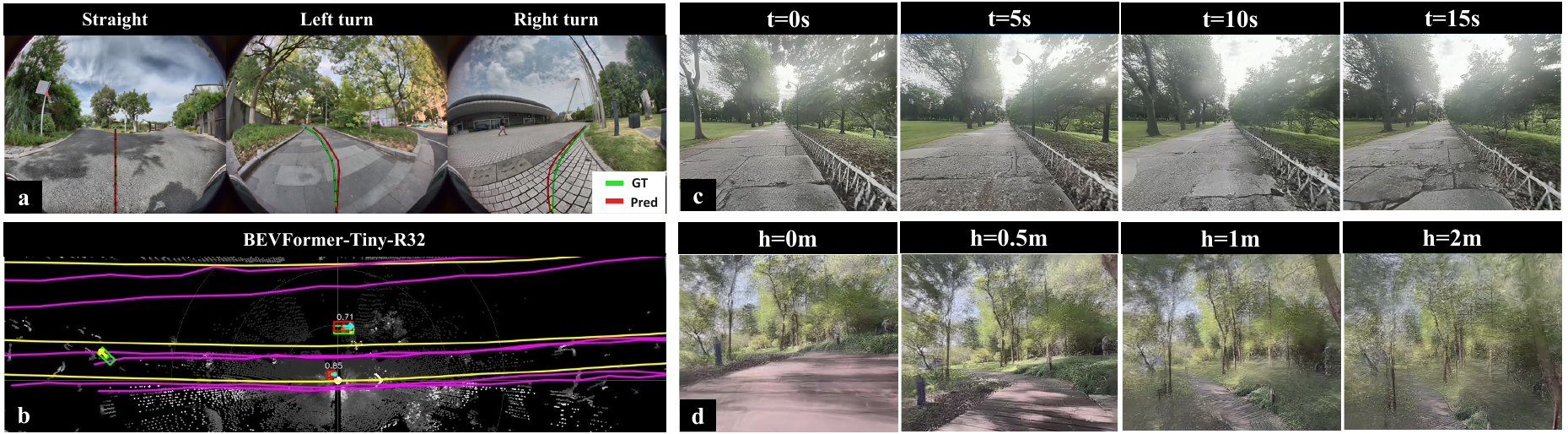}
    \caption{
    \textbf{Qualitative visualizations of the four benchmark tasks in \benchname{}.}
    \textbf{(a) Instruction-conditioned trajectory prediction:}
    examples of predicted future trajectories under three navigation commands.
    \textbf{(b) BEV-based perception of 3D objects and road structures.}
    \textbf{(c) Future scene prediction:}
    generated future egocentric observations at different rollout times.
    \textbf{(d) Multi-view scene reconstruction:}
    novel-view rendering results under different viewpoint changes.
    }
    \label{fig:four_task_visualization}
\end{figure*}

\subsection{Trajectory-Prediction Baselines and Results}
\textbf{Baselines.}
We evaluate five VLN models:
InternNav~\citep{internnav},
NaVILA~\citep{navila},
OmniNav~\citep{omninav},
StreamVLN~\citep{streamvln},
and SocialNav (Qwen3-VL-4B)~\citep{socialnav};
and five VLA models:
LingBot-VLA-v2~\citep{lingbotvla},
DepthVLA~\citep{depthvla},
OpenPI $\pi$0.5~\citep{pi05},
OpenVLA~\citep{openvla},
and Isaac-GR00T-N1.7~\citep{gr00tn1}.
All models are evaluated under the common trajectory-prediction protocol;
model-specific adaptations are detailed in
Appendix~\ref{sec:appendix_baselines}.

\textbf{Results.}
Table~\ref{tab:trajectory_results} reports the trajectory-prediction
results.
Among learned policies with pseudo-closed-loop evaluation,
NaVILA achieves the highest QNS of 0.843, followed by StreamVLN at 0.833.
For imitation accuracy, Isaac-GR00T-N1.7 obtains the lowest ADE (0.027),
while NaVILA achieves the lowest FDE, AHE, and FHE.
The constant-velocity and stationary references further show that
collision avoidance alone is insufficient, as useful navigation also
requires progress and instruction compliance.

\begin{table}[!htbp]
\centering
\caption{Instruction-conditioned trajectory prediction.}
\label{tab:trajectory_results}
\deeptablesetup
\setlength{\tabcolsep}{2pt}
\begin{tabularx}{\linewidth}{@{}L{0.25\linewidth}*{11}{C}@{}}
\toprule
& \multicolumn{4}{c}{\textbf{Trajectory imitation}}
& \multicolumn{6}{c}{\textbf{Replay-based navigation quality}}
& \textbf{Overall} \\
\cmidrule(lr){2-5}\cmidrule(lr){6-11}\cmidrule(l){12-12}
\textbf{Model}
& \deeptablehead{ADE}{$\downarrow$}
& \deeptablehead{FDE}{$\downarrow$}
& \deeptablehead{AHE}{$\downarrow$}
& \deeptablehead{FHE}{$\downarrow$}
& \deeptablehead{NC}{$\uparrow$}
& \deeptablehead{WAC}{$\uparrow$}
& \deeptablehead{TTC}{$\uparrow$}
& \deeptablehead{CL}{$\uparrow$}
& \deeptablehead{EP}{$\uparrow$}
& \deeptablehead{IC}{$\uparrow$}
& \deeptablehead{QNS}{$\uparrow$} \\
\midrule
\rowcolor{DeepTableBand}
\multicolumn{12}{@{}l@{}}{\textit{VLN-based policies}} \\

InternNav
& 0.079 & 0.098 & 0.038 & 0.062
& 0.945 & 0.955 & 0.871 & 0.815 & 0.893 & 0.818 & 0.764 \\

NaVILA
& \underline{0.030}
& \textbf{0.059}
& \textbf{0.028}
& \textbf{0.048}
& \textbf{0.968}
& \textbf{0.975}
& \textbf{0.912}
& \underline{0.838}
& \textbf{0.941}
& \textbf{0.893}
& \textbf{0.843} \\

OmniNav
& 0.090 & 0.152 & 0.033 & 0.054
& 0.939 & 0.949 & 0.863 & 0.811 & 0.866 & 0.803 & 0.743 \\

StreamVLN
& 0.036 & 0.067 & 0.032 & 0.053
& \underline{0.964}
& \underline{0.972}
& \underline{0.905}
& \textbf{0.843}
& \underline{0.933}
& \underline{0.884}
& \underline{0.833} \\

SocialNav Qwen3-VL-4B
& 0.295 & 0.538 & 0.061 & 0.106
& 0.843 & 0.862 & 0.811 & 0.825 & 0.851 & 0.725 & 0.757 \\

\midrule
\rowcolor{DeepTableBand}
\multicolumn{12}{@{}l@{}}{\textit{VLA-based policies}} \\

LingBot-VLA-v2
& 0.044 & 0.068 & 0.038 & 0.054
& 0.958 & 0.966 & 0.893 & 0.829 & 0.926 & 0.863 & 0.809 \\

DepthVLA
& 0.056 & 0.074 & 0.032 & \underline{0.049}
& 0.952 & 0.961 & 0.882 & 0.825 & 0.915 & 0.842 & 0.789 \\

OpenPI $\pi$0.5
& 0.036 & 0.075 & 0.032 & 0.052
& 0.961 & 0.969 & 0.899 & 0.834 & 0.922 & 0.874 & 0.819 \\

OpenVLA
& 0.041 & 0.068 & 0.030 & 0.051
& 0.956 & 0.964 & 0.888 & 0.823 & 0.928 & 0.855 & 0.802 \\

Isaac-GR00T-N1.7
& \textbf{0.027}
& \underline{0.062}
& \underline{0.029}
& 0.051
& 0.936 & 0.967 & 0.869 & 0.834 & 0.875 & 0.891 & 0.820 \\

\midrule
\rowcolor{DeepTableBand}
\multicolumn{12}{@{}l@{}}{\textit{Reference trajectories}} \\

Human trajectory
& 0 & 0 & 0 & 0
& 1.000 & 1.000 & 1.000 & 0.851 & 1.000 & 0.978 & 0.946 \\

Constant velocity
& 0.112 & 0.213 & 0.041 & 0.079
& 0.942 & 0.951 & 0.858 & 0.806 & 0.884 & 0.672 & 0.716 \\

Stationary
& 0.561 & 1.032 & 0.052 & 0.101
& 1.000 & 1.000 & 1.000 & 0.496 & 0.017 & 0.152 & 0.149 \\
\bottomrule
\end{tabularx}
\deeptablenote{Best and second-best results are shown in \textbf{bold} and \underline{underlined}, respectively.}
\end{table}

\subsection{BEV-Perception Baselines and Results}
\label{subsec:bev_protocol}

\textbf{Baselines.}
We evaluate BEVFormer-Tiny~\citep{li2022bevformer},
MapTRv2~\citep{MapTRv2},
BEVFusion~\citep{liu2022bevfusion},
and SparseDrive~\citep{sun2025sparsedrive}.
Their sensor inputs, training configurations, and checkpoint-selection
protocols are detailed in Appendix~\ref{sec:appendix_bev}.

\textbf{Results.}
Table~\ref{tab:bev_results} shows that SparseDrive achieves the
highest road-structure mAP, whereas BEVFusion obtains the highest box mAP
and HDS.
The latter result should be interpreted together with BEVFusion's
additional LiDAR input.
Additional qualitative results are provided in
Appendix~\ref{sec:appendix_bev}.

\begin{table}[!htbp]
\centering
\caption{BEV-based perception of road structures and 3D objects.}
\label{tab:bev_results}
\deeptablesetup
\begin{tabularx}{\linewidth}{@{}L{0.29\linewidth}L{0.25\linewidth}*{3}{C}@{}}
\toprule
\textbf{Method} & \textbf{Input}
& \deeptablehead{Line mAP}{$\uparrow$}
& \deeptablehead{Box mAP}{$\uparrow$}
& \deeptablehead{HDS}{$\uparrow$} \\
\midrule
BEVFormer-Tiny$^\dagger$
& Cam.+Pose
& 16.92
& 52.57
& 54.39 \\

MapTRv2$^\dagger$
& Camera
& 13.63
& 45.90
& 47.57 \\

BEVFusion
& Cam.+LiDAR
& 19.29
& \textbf{67.11}
& \textbf{64.48} \\

SparseDrive
& Cam.+Pose
& \textbf{21.08}
& 58.70
& 61.12 \\
\bottomrule
\end{tabularx}
\deeptablenote{Best and second-best results are shown in \textbf{bold} and \underline{underlined}, respectively. $^\dagger$ denotes early-stopped training. Aggregate scores are reported as percentages.}
\end{table}

\subsection{Future-Scene-Prediction Baselines and Results}
\label{sec:world_model}

\textbf{Baselines.}
We evaluate six representative world and video-generation models:
Matrix-Game-3~\citep{matrixgame3},
WorldStereo~\citep{worldstereo},
Cosmos3-Nano~\citep{cosmos3},
LingBot-World~\citep{lingbotworld},
Wan2.2-TI2V-5B~\citep{wan2025},
and CogVideoX1.5-5B-I2V~\citep{cogvideox}.
All methods are evaluated on the same 200-clip subset;
model details and temporal-normalization protocols are
provided in Appendix~\ref{sec:appendix_world_model}.

\textbf{Results.}
Table~\ref{tab:future_scene_results} shows that no single model dominates
all evaluation dimensions.
LingBot-World achieves the highest Q-Align-I, Q-Align-V, and LAION-Aes
scores; Cosmos3-Nano performs best on CLIP-I and CLIP-IQA+;
Wan2.2-TI2V-5B obtains the highest DOVER-T; and WorldStereo achieves the
lowest FVD.
The results reveal trade-offs among perceptual quality, temporal quality,
appearance consistency, and distributional realism.

\begin{table}[!htbp]
\centering
\caption{Future scene prediction.}
\label{tab:future_scene_results}
\deeptablesetup
\setlength{\tabcolsep}{2pt}
\begin{tabularx}{\linewidth}{@{}L{0.25\linewidth}*{9}{C}@{}}
\toprule
\textbf{Model}
& \deeptablehead{Q-Align}{I$\uparrow$}
& \deeptablehead{Q-Align}{V$\uparrow$}
& \deeptablehead{CLIP}{I$\uparrow$}
& \deeptablehead{CLIP}{IQA+$\uparrow$}
& \deeptablehead{LAION}{Aes$\uparrow$}
& \deeptablehead{DOVER}{T$\uparrow$}
& \deeptablehead{Smooth.}{$\uparrow$}
& \deeptablehead{Dynamic}{$\uparrow$}
& \deeptablehead{FVD}{$\downarrow$} \\
\midrule
Matrix-Game-3
& 2.759
& 0.474
& 76.72
& 0.380
& 4.374
& 0.066
& \textbf{0.978}
& \textbf{1.000}
& 964.1 \\

WorldStereo
& 3.310
& 0.633
& 85.73
& 0.474
& \underline{4.786}
& 0.050
& 0.896
& \underline{0.995}
& \textbf{386.2} \\

Cosmos3-Nano
& 3.340
& 0.619
& \textbf{87.85}
& \textbf{0.588}
& 4.770
& \underline{0.088}
& \underline{0.969}
& 0.895
& \underline{411.3} \\

LingBot-World
& \textbf{3.629}
& \textbf{0.717}
& 83.11
& \underline{0.575}
& \textbf{4.853}
& 0.078
& 0.942
& \textbf{1.000}
& 764.7 \\

Wan2.2-TI2V-5B
& \underline{3.445}
& \underline{0.660}
& \underline{87.73}
& 0.475
& 4.769
& \textbf{0.094}
& 0.955
& \textbf{1.000}
& 616.3 \\

CogVideoX1.5-5B-I2V
& 2.734
& 0.479
& 85.79
& 0.386
& 4.629
& 0.051
& 0.921
& 0.985
& 692.7 \\
\bottomrule
\end{tabularx}
\deeptablenote{Best and second-best results are shown in \textbf{bold} and \underline{underlined}, respectively.}
\end{table}

\subsection{Multi-View-Reconstruction Baselines and Results}
\label{sec:gs_experiment}

\textbf{Baselines.}
We evaluate five pretrained Gaussian-splatting methods without additional
fine-tuning:
ReCoSplat~\citep{cheng2026recosplat},
YoNoSplat~\citep{ye2026yonosplat},
AnySplat~\citep{jiang2025anysplat},
FreeSplatter-S~\citep{xu2025freesplatter},
and MVSplat~\citep{chen2024mvsplat}.
Model-specific reconstruction interfaces and evaluation settings are
provided in Appendix~\ref{sec:appendix_gs}.

\textbf{Results.}
Table~\ref{tab:reconstruction_results} shows complementary strengths across
methods.
ReCoSplat achieves the best PSNR, LPIPS, and MS-SSIM,
YoNoSplat obtains the highest SSIM, and AnySplat performs best on DISTS,
FSIMc, and GMSD.
No single method dominates all reconstruction metrics.

\begin{table}[!htbp]
\centering
\caption{Multi-view scene reconstruction.}
\label{tab:reconstruction_results}
\deeptablesetup
\begin{tabularx}{\linewidth}{@{}L{0.21\linewidth}*{8}{C}@{}}
\toprule
\textbf{Method} & \textbf{$K$}
& \deeptablehead{PSNR}{$\uparrow$}
& \deeptablehead{SSIM}{$\uparrow$}
& \deeptablehead{LPIPS}{$\downarrow$}
& \deeptablehead{MS-SSIM}{$\uparrow$}
& \deeptablehead{DISTS}{$\downarrow$}
& \deeptablehead{FSIMc}{$\uparrow$}
& \deeptablehead{GMSD}{$\downarrow$} \\
\midrule
ReCoSplat
& 8--16
& \textbf{15.937}
& \underline{0.2425}
& \textbf{0.5113}
& \textbf{0.3901}
& \underline{0.2133}
& \underline{0.6364}
& \underline{0.2436} \\

YoNoSplat
& 8--16
& \underline{15.390}
& \textbf{0.2736}
& 0.6721
& 0.3096
& 0.3986
& 0.5113
& 0.2928 \\

AnySplat
& 8--16
& 14.544
& 0.2096
& \underline{0.5119}
& \underline{0.3359}
& \textbf{0.2128}
& \textbf{0.6429}
& \textbf{0.2393} \\

FreeSplatter-S
& 2
& 14.537
& 0.2373
& 0.6461
& 0.3099
& 0.3277
& 0.5687
& 0.2656 \\

MVSplat
& 2
& 11.777
& 0.1938
& 0.6307
& 0.3008
& 0.3312
& 0.5807
& 0.2774 \\
\bottomrule
\end{tabularx}
\deeptablenote{Best and second-best results are shown in \textbf{bold} and \underline{underlined}, respectively. $K$ is the number of context views.}
\end{table}

Figure~\ref{fig:four_task_visualization} provides representative qualitative
examples across the four benchmark tasks.
More extensive qualitative results for trajectory prediction, BEV perception,
future scene prediction, and multi-view scene reconstruction are provided in
Appendices~D, E, F, and G, respectively.

\section{Conclusion}
\label{sec:conclusion}

We present \benchname{}, a large-scale real-world benchmark for outdoor
wheel-legged robot autonomy that combines synchronized multimodal sensing,
dense spatial annotations, behavior-level navigation instructions, and
continuous future trajectories.
Built on the same real-world robot data, the benchmark supports four
complementary tasks: instruction-conditioned trajectory prediction,
BEV-based perception of 3D objects and road structures, future scene
prediction, and multi-view scene reconstruction.
For trajectory prediction, we further introduce a log-replay-based
pseudo-closed-loop evaluator that complements imitation metrics with
safety, progress, walkable-area compliance, and instruction adherence.

Across representative VLN/VLA policies, BEV perception models, world
models, and reconstruction methods, our experiments reveal substantial
remaining challenges in close-range interaction, spatial understanding,
future-scene modeling, and viewpoint-dependent reconstruction.
These results suggest that outdoor wheel-legged robot autonomy remains
difficult even when perception, navigation, prediction, and reconstruction
are evaluated on the same embodiment and scene distribution.
We hope \benchname{} provides a common real-world foundation for studying
these capabilities together and supports future progress toward more
integrated and socially compliant outdoor robot autonomy.

\clearpage
\bibliographystyle{plainnat}
\begingroup
\small
\setlength{\bibsep}{3pt}
\bibliography{references}
\endgroup

\clearpage
\appendix
\section{Additional Related Work}
\label{sec:additional_related_work}

\subsection{Real-World Multimodal Robot Datasets}
\label{app:robot_datasets}

Existing real-world robot datasets provide complementary forms of
multimodal sensing and navigation supervision.
M2DGR~\citep{m2dgr} provides synchronized fisheye cameras, LiDAR,
inertial sensing, GNSS, and odometry for multi-sensor SLAM, while
M3ED~\citep{m3ed} extends multimodal sensing to multiple robotic
platforms and challenging environments.
GrandTour~\citep{grandtour} further provides synchronized multimodal
observations, proprioception, and high-precision trajectories for
legged-robot perception and state estimation.

Several datasets focus more directly on navigation.
RECON~\citep{recon} provides real-robot trajectories for
goal-conditioned open-world navigation, while
SCAND~\citep{scand} collects human-teleoperated demonstrations for
socially compliant navigation.
GND~\citep{gnd} introduces semantic traversability supervision for
outdoor global navigation, and DiTer~\citep{diter} combines multimodal
quadruped observations with reference trajectories and global maps
across diverse outdoor terrains.

These datasets emphasize different subsets of sensing, localization,
trajectory, or traversability supervision.
\benchname{} instead integrates temporally aligned multimodal sensing,
persistent 3D objects, road-structure geometry, structured navigation
instructions, and future trajectories in a single wheel-legged robot
dataset, allowing the same recordings to support multiple perception,
navigation, prediction, and reconstruction tasks.

\subsection{Navigation Policies and Planning Evaluation}
\label{app:navigation_related}

Recent navigation models increasingly map multimodal observations and
semantic instructions directly to continuous robot behavior.
NaVILA~\citep{navila} adapts vision-language-action modeling to legged
robot navigation, while StreamVLN~\citep{streamvln} introduces
streaming visual-language reasoning for continuous navigation.
InternNav~\citep{internnav} further explores generalizable
vision-language navigation with a dual-system architecture.

General-purpose VLA models provide another relevant policy family.
OpenVLA~\citep{openvla} learns continuous robot actions from
vision-language inputs, while $\pi$0.5~\citep{pi05} uses a
flow-matching action expert to generate continuous action chunks.
These approaches are originally developed under different embodiments,
action spaces, and task definitions.
In our benchmark, they are adapted to a common observation, instruction,
robot-state, and continuous-trajectory interface, enabling direct
comparison under the same real-world navigation data.

Planning evaluation also increasingly extends beyond open-loop
trajectory displacement.
NAVSIM~\citep{navsim} evaluates predicted driving trajectories through
non-reactive scene replay and combines multiple safety- and
progress-related criteria.
Our pseudo-closed-loop evaluator follows this general replay-based
principle but adapts it to wheel-legged robot navigation, where nearby
pedestrians and riders, curb and sidewalk boundaries, and
behavior-level instructions directly constrain acceptable trajectories.

\subsection{BEV Perception for Driving and Robot Navigation}
\label{app:bev_related}

Large-scale autonomous-driving datasets including
nuScenes~\citep{nuscenes}, Waymo Open Dataset~\citep{waymo}, and
Argoverse~2~\citep{argoverse2} have established widely used benchmarks
for 3D object detection and spatial scene understanding.

BEV-based perception methods provide a natural representation for
combining surrounding observations and geometric reasoning.
BEVFormer~\citep{li2022bevformer} constructs spatiotemporal BEV
representations from multi-camera images.
BEVFusion~\citep{liu2022bevfusion} combines camera and LiDAR
information in a unified BEV representation, while
MapTRv2~\citep{MapTRv2} represents road and map elements using
vectorized geometric predictions.

These approaches motivate the perception baselines used in our
benchmark, but the spatial constraints differ from conventional
autonomous driving.
Vehicle-oriented benchmarks primarily emphasize traffic participants,
lanes, and HD-map elements, whereas outdoor wheel-legged robots
frequently operate near pedestrians, riders, curbs, sidewalk
boundaries, road--sidewalk dividing lines, and crosswalks.
Our BEV task therefore evaluates both nearby 3D objects and
navigation-relevant road structures in the robot-centric coordinate
frame.

\subsection{World Models for Future Scene Prediction}
\label{app:world_model_related}

World models predict how future observations evolve under actions,
motion conditions, or other control signals.
GAIA-1~\citep{gaia1} demonstrates generative world modeling for
autonomous driving, while Navigation World Models~\citep{nwm}
generate future egocentric observations conditioned on navigation
actions.
Cosmos~\citep{cosmos} further explores large-scale world foundation
models for physical-AI applications.

These works demonstrate the potential of generative prediction for
reasoning about future scene evolution, but are commonly evaluated in
autonomous-driving, generic video-generation, or other embodiment
settings.
Outdoor wheel-legged robot navigation introduces a distinct visual
regime, including low camera height, close-range pedestrians and
riders, vegetation, curbs, and frequent viewpoint changes.
Our benchmark evaluates whether existing world models can generate
future robot-centric observations under these conditions using
perceptual, temporal, conditioning-consistency, and distributional
metrics.

\subsection{Neural Scene Reconstruction and Scenario Replay}
\label{app:reconstruction_related}

3D Gaussian Splatting~\citep{3dgs} provides an efficient explicit
scene representation for novel-view synthesis.
Recent methods extend Gaussian-based reconstruction to sparse,
unposed, and online observations.
MVSplat~\citep{chen2024mvsplat} uses posed multi-view cost volumes,
FreeSplatter~\citep{xu2025freesplatter} targets sparse-view pose-free
reconstruction, and AnySplat~\citep{jiang2025anysplat} reconstructs
Gaussian scene representations from unconstrained views.
ReCoSplat~\citep{cheng2026recosplat} performs online reconstruction
through render-and-compare, while
YoNoSplat~\citep{ye2026yonosplat} supports both posed and unposed
feed-forward reconstruction.

Beyond novel-view synthesis, reconstructed Gaussian scenes have also
been explored for scenario replay and simulation.
HUGSIM~\citep{hugsim}, for example, integrates Gaussian-splatting-based
scene reconstruction into a photorealistic closed-loop driving
simulator.
Our current reconstruction benchmark is more limited in scope: it
evaluates observation-grounded novel-view rendering on real robot
navigation sequences rather than claiming full closed-loop simulation
capability.
Together with the world-model task, it provides complementary
evaluation of future visual prediction and explicit scene
reconstruction.

\section{Dataset Details}
\label{app:dataset_details}

\subsection{Benchmark Task Pipeline}
\label{sec:appendix_pipeline}

\begin{figure}[h]
    \centering
    \includegraphics[width=\textwidth]{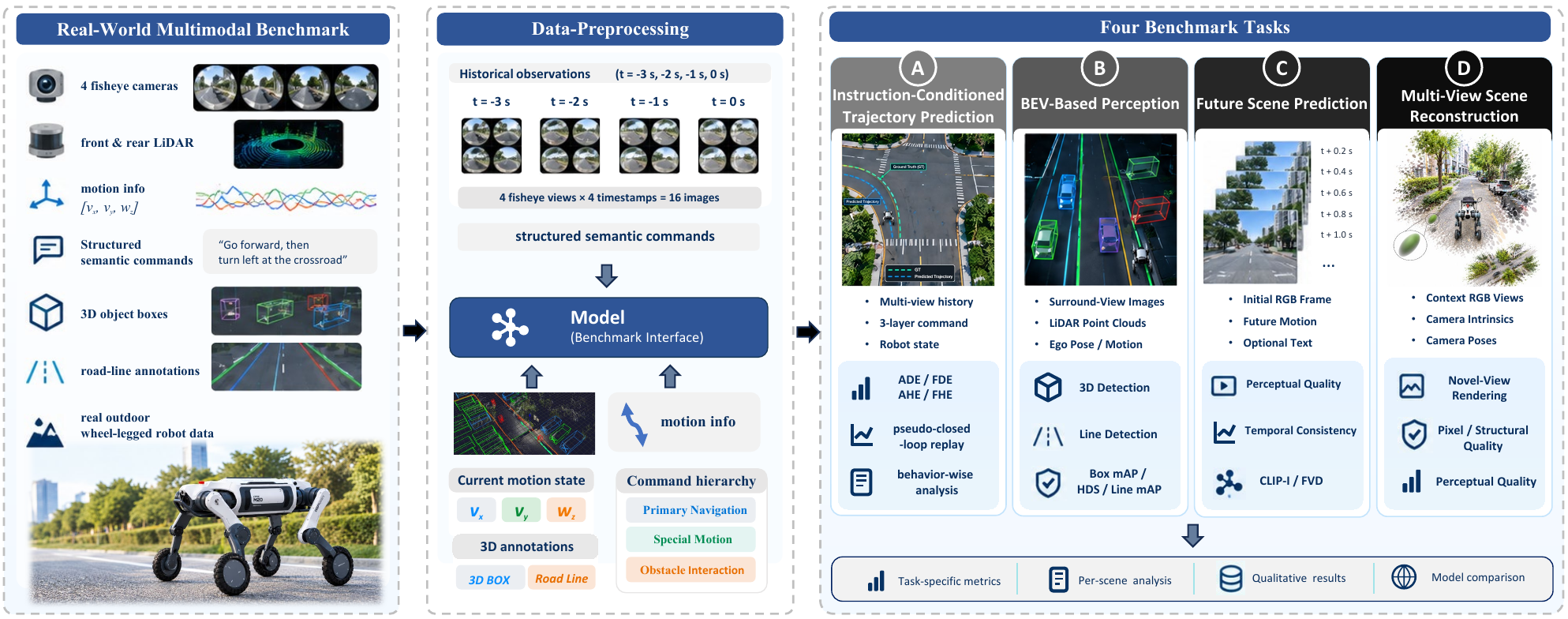}
    \caption{
\textbf{Benchmark task pipeline.}
The same real-world multimodal recordings and structured annotations
support four tasks:
(A) instruction-conditioned trajectory prediction,
(B) BEV-based perception of 3D objects and road structures,
(C) future scene prediction, and
(D) multi-view scene reconstruction.
Task-specific inputs are constructed from the shared recordings;
LiDAR, ego-motion, instruction, and camera-pose conditions are used when
required by the corresponding model.
Outputs are evaluated using task-specific quantitative metrics and
qualitative comparisons.
}
    \textbf{}
    \label{fig:pipeline}
\end{figure}

\subsection{Sensor Configuration and Calibration}
\label{app:sensor_configuration}

Table~\ref{tab:sensor_configuration} summarizes the sensing and state
modalities of the data-collection platform.

\begin{figure}[h]
    \centering
    \includegraphics[width=\textwidth]{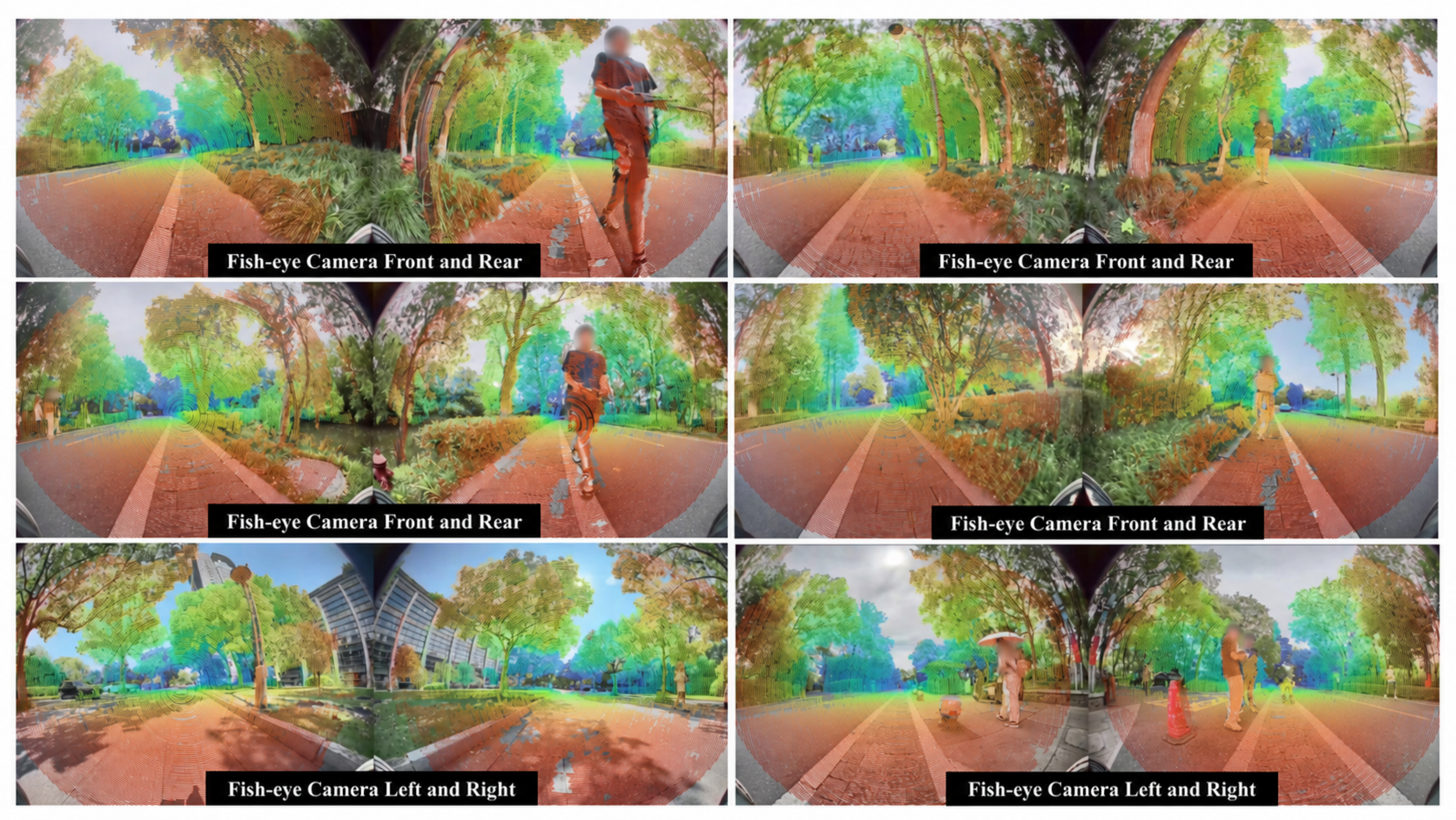}
    \caption{
\textbf{Camera--LiDAR calibration validation.}
LiDAR point clouds are projected onto the calibrated fisheye camera
views to qualitatively verify cross-modal alignment.
Projected points are color-coded by depth along the camera axis,
with red indicating nearer points and blue indicating farther points.
}
    \textbf{}
    \label{fig:calibration}
\end{figure}

\textbf{Calibration.}
The four fisheye cameras are individually calibrated for camera intrinsics,
fisheye distortion, and extrinsics with respect to the robot body frame.
The front and rear LiDARs are registered to the same robot-centric frame.
As shown in Fig.~\ref{fig:calibration}, cross-modal calibration is
qualitatively verified by projecting LiDAR point clouds onto the calibrated
fisheye images.
Projected points are color-coded by depth along the camera axis, with red
indicating nearer points and blue indicating farther points.

\textbf{Temporal synchronization.}
Annotated frames are sampled at 1~Hz from the 10~Hz camera and LiDAR
recordings.
For each annotation timestamp, the temporally closest camera images and
LiDAR sweeps are selected, while LIO and robot-state measurements are
aligned to the same anchor timestamp.
The resulting anchor therefore provides temporally aligned visual,
geometric, ego-motion, and robot-state observations for downstream
annotation and benchmark construction.
\begin{table}[t]
    \centering
    \caption{Sensing and state modalities of the data-collection
    platform.}
    \label{tab:sensor_configuration}
    \renewcommand{\arraystretch}{1.08}
    \resizebox{0.8\columnwidth}{!}{
    \begin{tabular}{lcccc}
        \toprule
        \textbf{Modality}
        & \textbf{Number}
        & \textbf{Specification}
        & \textbf{Coverage}
        & \textbf{Native Rate} \\
        \midrule

        Fisheye Camera
        & 4
        & $1920\times1536$
        & Front/Rear/Left/Right
        & 10\,Hz \\

        3D LiDAR
        & 2
        & RoboSense Airy, 96 ch.
        & Front/Rear
        & 10\,Hz \\

        IMU
        & 1
        & Inertial measurements
        & Ego state
        & 200\,Hz \\

        LIO/Odometry
        & 1
        & 6-DoF pose
        & Ego motion
        & $\sim$200\,Hz \\

        Motion State
        & 1
        & $v_x$, $v_y$, $\omega_z$
        & Ego motion
        & $\sim$20\,Hz \\

        GPS
        & 1
        & Global positioning
        & Reference only
        & $\sim$2\,Hz \\

        \bottomrule
    \end{tabular}
    }
\end{table}

\subsection{Trajectory Construction Details}
\label{app:trajectory_construction}

Reference future trajectory supervision is constructed from the recorded
motion-state stream.
At each time step, the motion state provides the longitudinal velocity
$v_x$, lateral velocity $v_y$, and yaw rate $\omega_z$ in the
robot-centric FLU coordinate system.

Given the body-frame velocities
$(v_x^k,v_y^k,\omega_z^k)$ at sampling interval $\Delta t$, the planar
pose $(x_k,y_k,\psi_k)$ is integrated according to

\begin{equation}
\begin{aligned}
x_{k+1}
&=
x_k+
\left(
v_x^k\cos\psi_k-
v_y^k\sin\psi_k
\right)\Delta t,
\\
y_{k+1}
&=
y_k+
\left(
v_x^k\sin\psi_k+
v_y^k\cos\psi_k
\right)\Delta t,
\\
\psi_{k+1}
&=
\psi_k+\omega_z^k\Delta t .
\end{aligned}
\label{eq:motion_integration}
\end{equation}

The integrated sequence is resampled to 10\,Hz and transformed into
the robot coordinate frame at the prediction anchor.
Each annotated frame is therefore paired with the subsequent robot
motion represented as a sequence of planar poses.
LIO trajectories are retained separately for tasks requiring metric
camera motion.

\label{sec:appendix_trajectory}

\subsection{Annotation Taxonomy}
\label{app:annotation_taxonomy}
\begin{table}[htbp]
    \centering
    \caption{Annotation taxonomy: 27 object categories, 4
    road-structure classes, and 20 navigation instructions in three layers.}
    \label{tab:annotation_taxonomy}
    \renewcommand{\arraystretch}{1.1}
    \resizebox{0.8\columnwidth}{!}{
    \begin{tabular}{l>{\raggedright\arraybackslash}p{11.5cm}}
        \toprule
        \textbf{Group} & \textbf{Categories} \\
        \midrule
        \multicolumn{2}{l}{\textit{3D objects (27)}} \\
        Human
        & Pedestrian, Pedestrian Accessories \\
        Mobility
        & Bicycle, Bicycle Rider, Scooter, Scooter Rider,
          Motorcycle, Motorcycle Rider, Tricycle, Tricycle Rider \\
        Vehicle
        & Car, SUV, Van, Truck, Bus \\
        Traffic Facility
        & Cone, Traffic Barrier, Road Barrel,
          Reflective Triangle, Fire Hydrant \\
        Other
        & Robot Dog, Platform Cart, Construction Cart,
          Excavator, Forklift, Dog, Cat \\
        \midrule
        \multicolumn{2}{l}{\textit{Road structure Line (4)}} \\

        & Sidewalk Curb, Road Curb, Road--Sidewalk Dividing Line,
          Crosswalk \\
        \midrule
        \multicolumn{2}{l}{\textit{Navigation instructions (20)}} \\
        Primary Navigation
        & \texttt{follow\_path},
          \texttt{continue\_straight\_at\_junction},
          \texttt{take\_left\_branch},
          \texttt{take\_right\_branch},
          \texttt{u\_turn},
          \texttt{unknown\_navigation} \\
        Special Motion
        & \texttt{stationary},
          \texttt{reversing},
          \texttt{rotate\_in\_place\_left/right},
          \texttt{lane\_left/right},
          \texttt{lateral\_left/right},
          \texttt{mixed\_motion} \\
        Obstacle Interaction
        & \texttt{obstacle\_wait},
          \texttt{obstacle\_avoid\_left/right},
          \texttt{obstacle\_reroute},
          \texttt{unknown\_obstacle\_interaction} \\
        \bottomrule
    \end{tabular}
    }
\end{table}

\subsection{Data Collection and Quality Control}
\label{app:data_quality_control}

\textbf{Socially compliant data collection.}
Data collection is performed by trained human operators following
predefined socially compliant navigation guidelines.
The protocol covers common interactions in shared outdoor spaces,
including yielding to pedestrians, maintaining safe clearance,
passing obstacles on an appropriate side, and negotiating encounters
with pedestrians, riders, and vehicles.
These guidelines are used to improve the behavioral consistency and
safety of the recorded expert trajectories.

\textbf{Annotation quality control.}
All annotations are created manually and pass through a dedicated review
loop.
Reviewers inspect annotated frames for completeness and consistency, and
frames containing missing, misaligned, or inconsistent labels are returned
for correction before acceptance.

\textbf{Demonstration quality control.}
Recorded expert demonstrations additionally undergo behavior-level review.
Human--robot interaction segments are inspected for consistency with the
predefined socially compliant navigation guidelines.
Segments containing unsafe, socially inappropriate, or inconsistent
operator behavior are manually identified and excluded from the benchmark.

\section{Navigation Evaluation Details}
\label{sec:appendix_navigation}

\textbf{Trajectory imitation metrics.}
Given predicted and reference trajectories, ADE and FDE measure the mean
and final Euclidean position errors, respectively.
AHE and FHE measure the corresponding mean and final absolute heading
errors, with angular differences wrapped to $[-\pi,\pi)$.

\textbf{Pseudo-closed-loop replay.}
The replay uses the predicted pose
$\hat{\mathbf p}_k=(\hat x_k,\hat y_k,\hat\psi_k)$ at 0.1-s intervals
and a rectangular ego footprint.
Annotated objects are replayed as BEV rectangles by interpolating box
centers and yaw between 1-Hz keyframes using persistent instance identities;
an object present in only one neighboring keyframe is held fixed.
Road-structure polylines from the anchor keyframe define the local
walkable-area constraints.
The replay is non-reactive.

\begin{itemize}[leftmargin=*,itemsep=1pt,topsep=2pt,parsep=0pt]

\item \textbf{No Collision (NC):}
0 for an at-fault collision with a person, rider, vehicle, animal, or
robot; 0.5 for an inanimate object; and 1 otherwise.
A collision is not at fault when the ego is stationary and another
object moves into it.

\item \textbf{Walkable-Area Compliance (WAC):}
0 if the ego footprint crosses a curb or a road--sidewalk dividing line
outside a crosswalk, unless the recorded trajectory crosses the same
boundary within the horizon; 1 otherwise.

\item \textbf{Time-to-Collision (TTC):}
0 if 1-s constant-velocity propagation from any predicted step intersects
a replayed agent; 1 otherwise.

\item \textbf{Clearance (CL):}
$\min(1,d_{\min}/0.5\,\mathrm{m})$, where $d_{\min}$ is the minimum
distance to any person, rider, or animal.

\item \textbf{Ego Progress (EP):}
predicted progress projected onto the recorded path, normalized by the
recorded-path length and clipped to $[0,1]$; set to 1 when the recorded
motion is below 0.2\,m.

\item \textbf{Instruction Compliance (IC):}
checks instruction-specific trajectory geometry: at least $20^\circ$
signed heading change for branches, U-turns, and in-place rotation;
signed dominant lateral velocity for lateral or lane moves;
negative longitudinal velocity for reversing;
displacement below 0.2\,m for stationary or waiting;
passage on the instructed side for obstacle avoidance;
and continued motion when the operator moved for path following or
continuing straight.

\end{itemize}

\section{Trajectory-Prediction Baseline Adaptation Details}
\label{sec:appendix_baselines}

All trajectory-prediction experiments use the 8:1:1 clip-level
train/validation/test split described in Sec.~\ref{subsec:dataset_statistics}.
Each model receives the same observation history, four-camera visual input,
current robot state, semantic navigation instruction, and future trajectory
target, and predicts the complete $10\times3$ Body-FLU trajectory over the
future 1\,s.
Ground-truth 3D object and road-structure annotations are used only by the
evaluator and are not provided as test-time model inputs.

InternNav and StreamVLN use four-layer Transformer trajectory decoders with
Flow Matching to jointly generate the complete trajectory.
NaVILA and OmniNav use two-layer Transformer decoders for parallel regression
of all $10\times3$ trajectory variables.
LingBot-VLA-v2 and OpenPI $\pi$0.5 retain their native Flow-Matching action
experts; DepthVLA retains its Flow-Matching action head adapted to the
$10\times3$ output; and OpenVLA uses a two-layer Transformer decoder for
parallel continuous trajectory regression.

SocialNav builds on Qwen3-VL-4B and augments its visual tokens with
camera-identity, temporal-slot, and continuous timestamp embeddings.
A four-layer ActionFormer and motion-conditioned Flow-Matching head then
generate the complete trajectory.

Isaac-GR00T-N1.7 replaces the original manipulation diffusion head with a
two-layer, eight-head Transformer trajectory decoder conditioned on learned
horizon queries and the current motion state.
The decoder predicts normalized residuals over a constant-twist prior for all
$10\times3$ trajectory variables.

\begin{figure}[t]
    \centering
    \includegraphics[width=0.8\textwidth]{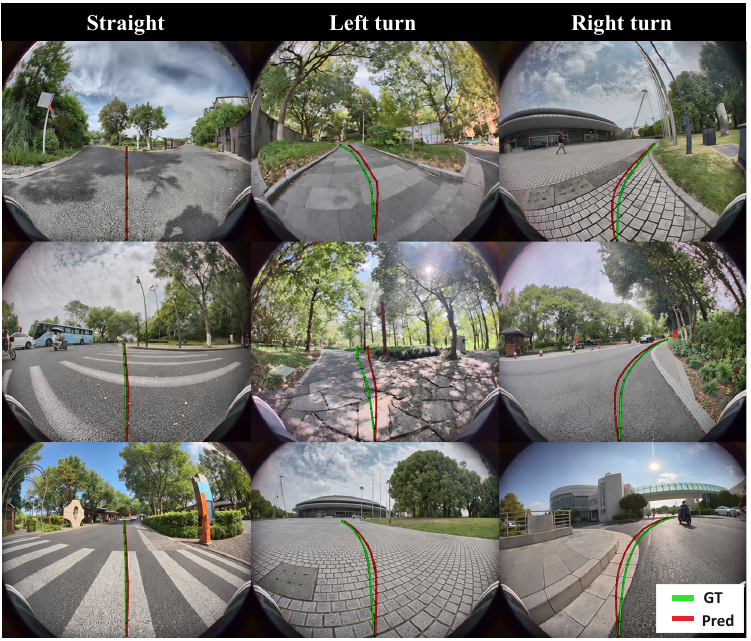}
\caption{
\textbf{Qualitative trajectory predictions with NaVILA.}
Columns correspond to three navigation instructions:
continue straight at junction, take left branch, and take right branch.
Red curves denote predicted trajectories, while green curves denote
ground-truth trajectories.
}

    \label{fig:vis}
\end{figure}

All future trajectory variables are predicted jointly rather than
waypoint-wise autoregressively.
Representative trajectory predictions are shown in
Fig.~\ref{fig:vis}.

\section{BEV-Based Perception Details}
\label{sec:appendix_bev}
\textbf{3D object metrics.}
Confidence-ranked predictions are matched one-to-one by ground-plane
center distance.
Box mAP averages AP over Car and Pedestrian and center-distance thresholds
of $\{0.5,1,2,4\}$\,m.
mATE, mASE, and mAOE use 2-m matching, with mASE defined as one minus
box-size IoU after center and yaw alignment.
We report
\begin{equation}
\mathrm{HDS}
=
\frac{
5\,\mathrm{mAP}
+
\sum_{k\in\{\mathrm{ATE,ASE,AOE}\}}
\max(0,1-m_k)
}{8},
\end{equation}
where $m_k$ denotes the corresponding detection error.
HDS is computed from unscaled values; the aggregate scores in
Table~\ref{tab:bev_results} are reported as percentages.

\textbf{Road-structure metrics.}
Road Curb, Sidewalk Curb, and Road--Sidewalk Dividing Line are merged into
the ordinary-line class, while Crosswalk forms the closed-contour class.
Predictions and references are clipped to
$[-30,30]\times[-15,15]$\,m, sampled along arc length, and matched
one-to-one using symmetric Chamfer distance.
Each class AP averages the $\{0.5,1,1.5\}$\,m matching thresholds,
and line-detection mAP is the equal-weight mean of ordinary-line and
crosswalk AP.

\textbf{Experimental settings.}
The BEV split contains 46,111 training, 5,917 validation, and 5,950 test
frames.
All four baselines use the same split, evaluator, and common optimization
schedule.
The adapted configurations use a ResNet-34/FPN initialized from the same
public ImageNet weights, an effective batch size of 192, and validation
ordinary-line AP@0.5\,m for checkpoint selection.
MapTRv2 is framewise and camera-only;
BEVFormer-Tiny and SparseDrive use cameras and ego pose for temporal
processing; BEVFusion uses cameras and LiDAR.
All reported runs use seed 42.
BEVFormer-Tiny and MapTRv2 were early-stopped after 92.7\% and 89.0\%
of their scheduled iterations, respectively.

\begin{figure}[htbp]
    \centering
    \includegraphics[width=\textwidth]{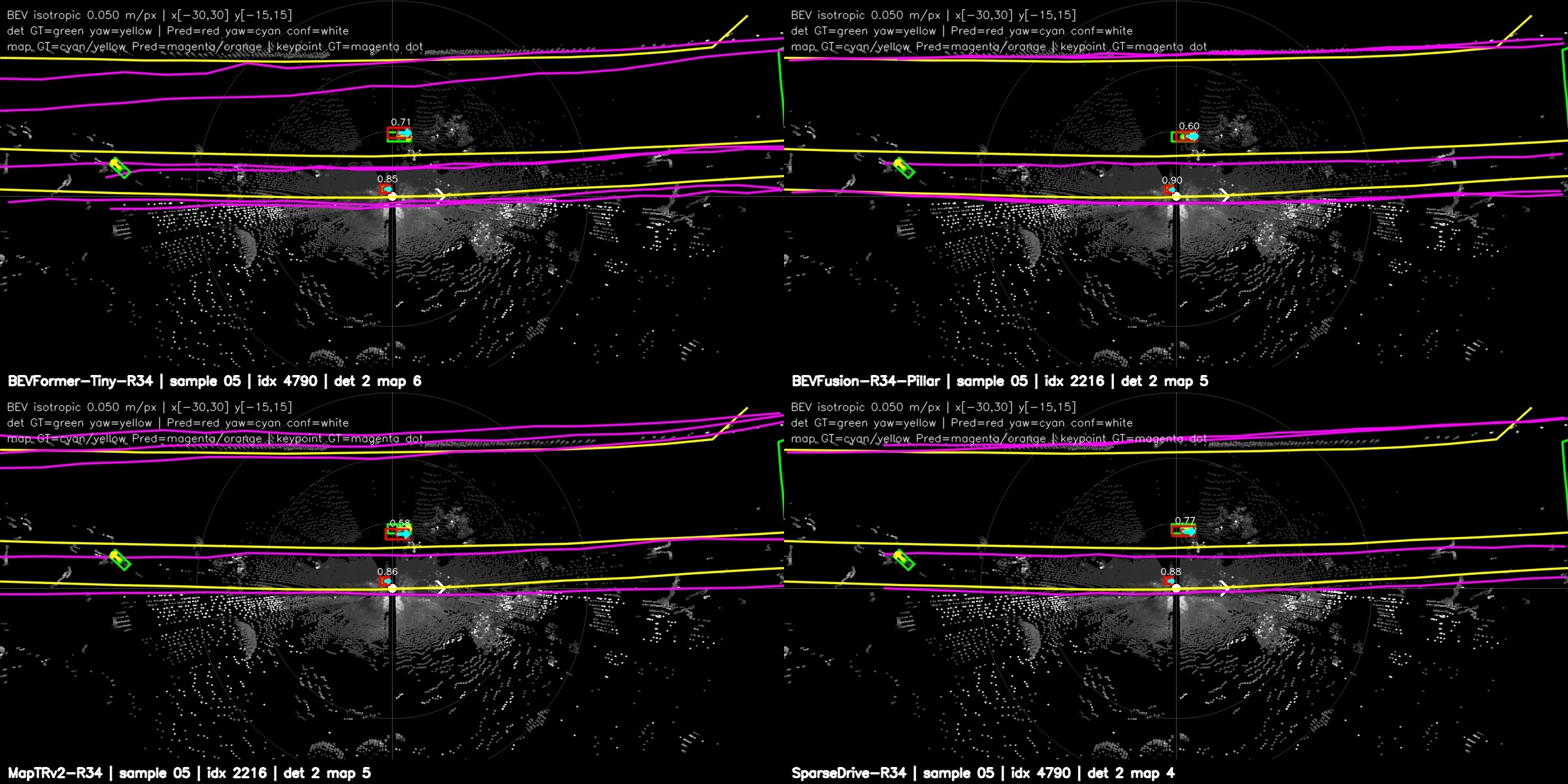}
\caption{Qualitative line detection comparison on a simple straight-road
test scene. The panels show BEVFormer-Tiny (top left), BEVFusion (top
right), MapTRv2 (bottom left), and SparseDrive (bottom right).
Ground-truth road geometry is shown in cyan, yellow, green, and blue;
predictions are shown in magenta and orange. Predictions with confidence
at least 0.25 are displayed. The candidate pool was constructed using
ground-truth content only.}
\label{fig:line_qualitative}
\end{figure}

Additional qualitative results are shown in
Fig.~\ref{fig:line_qualitative}.

\section{Future Scene Prediction Details}
\label{sec:appendix_world_model}

\subsection{Task Formulation and Conditioning}
\label{subsec:wm_task_details}

Given an initial rectified front-view image
$\mathbf{I}_{0}$, a future motion condition
$\mathbf{U}_{1:T}$, and an optional text condition
$\mathbf{c}$, a model generates a future egocentric RGB video:
\begin{equation}
    \hat{\mathbf{V}}_{1:T}
    =
    f_{\theta}
    \left(
        \mathbf{I}_{0},
        \mathbf{U}_{1:T},
        \mathbf{c}
    \right),
\end{equation}
where
$\hat{\mathbf{V}}_{1:T}
=
\{\hat{\mathbf{I}}_1,\ldots,\hat{\mathbf{I}}_T\}$.

For camera- or action-conditioned world models,
$\mathbf{U}_{1:T}$ contains the recorded future ego-motion after being
converted to the representation and temporal convention required by
the corresponding model.
For generic image-to-video models without an explicit motion-control
interface, $\mathbf{U}_{1:T}$ is omitted and a fixed forward-motion text
prompt is provided.
These methods therefore serve primarily as visual-generation baselines
rather than directly controllable world models.

\subsection{Evaluation Subset}
\label{subsec:wm_subset}

The evaluation subset contains 200 deterministic navigation clips
selected from the test split according to front-camera video and
trajectory synchronization requirements.
Each sample contains a rectified front-view pinhole video with a
horizontal field of view of $90^{\circ}$, the first RGB observation
frame, camera intrinsics, and synchronized camera and body poses.

\subsection{Model-Specific Conditioning Interfaces}
\label{subsec:wm_conditioning}

We preserve the native conditioning interface of each evaluated model.
Table~\ref{tab:world_model_protocol} summarizes the visual input,
motion or text condition, and native generation frame rate.

\begin{table}[t]
    \centering
    \caption{
    World-model baselines and their evaluation interfaces.
    $I_0$ denotes the first rectified front-view frame.
    Native frame rates and conditioning formats are retained during
    generation.
    }
    \label{tab:world_model_protocol}
    \small
    \resizebox{\columnwidth}{!}{
    \begin{tabular}{llclc}
        \toprule
        \textbf{Model}
        & \textbf{Model Type}
        & \textbf{Visual Input}
        & \textbf{Motion / Text Condition}
        & \textbf{FPS}
        \\
        \midrule

        Matrix-Game-3
        & Action-conditioned world model
        & $I_0$
        & Trajectory-derived keyboard/mouse actions
        & 17
        \\

        WorldStereo
        & Camera-controlled world model
        & $I_0$
        & Camera intrinsics + W2C trajectory
        & 16
        \\

        Cosmos3-Nano
        & Action-conditioned world model
        & $I_0$
        & Resampled 9-D motion sequence
        & 10
        \\

        LingBot-World
        & Camera-controlled world model
        & $I_0$
        & Interpolated camera trajectory
        & 16
        \\

        Wan2.2-TI2V-5B
        & Generic image-to-video
        & $I_0$
        & Forward-motion text prompt
        & 24
        \\

        CogVideoX1.5-5B-I2V
        & Generic image-to-video
        & $I_0$
        & Forward-motion text prompt
        & 16
        \\

        \bottomrule
    \end{tabular}
    }
\end{table}

\subsection{Temporal Normalization}
\label{subsec:wm_temporal}

The evaluated models generate videos with different native frame rates,
resolutions, and rollout lengths.
To ensure comparable evaluation, each generated video is divided into
16 equal temporal intervals, and the midpoint frame of each interval is
deterministically selected.

Q-Align-I, CLIP-I, CLIP-IQA+, and LAION-Aes are averaged over the sampled
frames.
FVD uses the same 16-frame sampling procedure for both generated and real
videos.
Q-Align-V, DOVER-T, Motion Smoothness, and Dynamic Degree follow the
temporal sampling procedures of their official implementations.

\subsection{World-Model Metric Definitions}
\label{subsec:wm_metrics}

We evaluate four complementary aspects of generated-video quality.

\textbf{Perceptual quality.}
Q-Align-I~\citep{qalign} evaluates frame-level perceptual quality.
CLIP-IQA+~\citep{clipiqa} provides a semantic-aware no-reference
image-quality score, while LAION-Aes~\citep{laion_aes} measures visual
aesthetic quality.

\textbf{Temporal quality.}
Q-Align-V~\citep{qalign} evaluates overall video quality and temporal
coherence.
DOVER-T~\citep{dover} measures technical video quality, while Motion
Smoothness and Dynamic Degree~\citep{vbench} characterize temporal
continuity and the amount of temporal variation.

\textbf{Appearance consistency.}
CLIP-I measures appearance consistency between generated frames and the
initial observation.
For each sampled generated frame, cosine similarity with the first frame
of the corresponding real video is computed using an OpenCLIP ViT-B/32
image encoder~\citep{openclip} and multiplied by 100.

\textbf{Distributional realism.}
Fr\'{e}chet Video Distance (FVD)~\citep{fvd} measures the distance between
generated and real video distributions using I3D video features.
FVD is computed jointly over all real and generated videos for each
evaluated model; lower values indicate a generated distribution closer to
the real-world robot observations.

\subsection{Benchmark Scope}

The current evaluation primarily measures visual fidelity, temporal
quality, appearance consistency, and distributional realism.
It does not directly evaluate whether the generated motion exactly follows
the intended future motion condition.
Explicit motion-conditioning accuracy therefore remains an important
extension of the benchmark.

\section{Multi-View Scene Reconstruction Details}
\label{sec:appendix_gs}

\label{sec:appendix_3dgs}
\begin{figure}[htbp]
    \centering
    \includegraphics[width=0.75\textwidth]{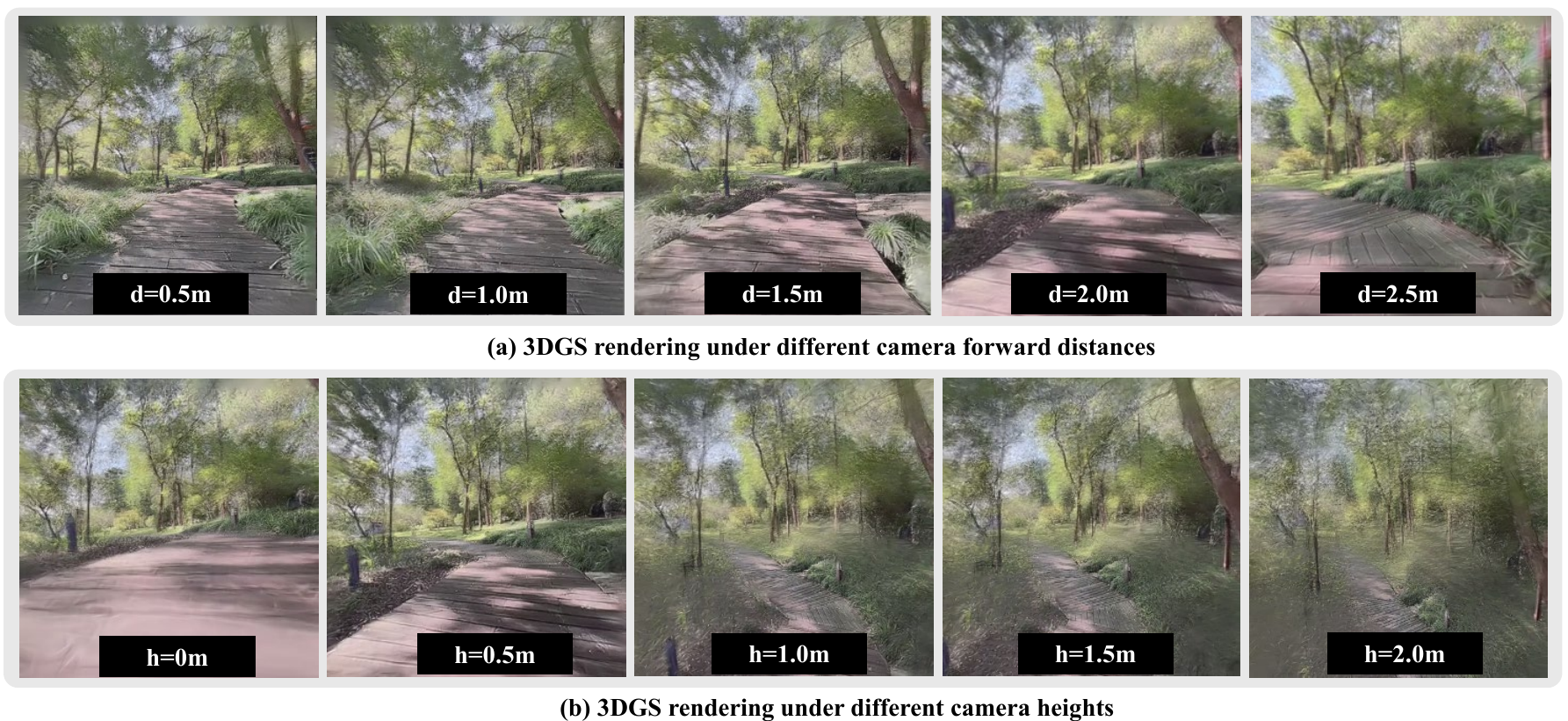}
\caption{
\textbf{Qualitative 3DGS multi-view scene reconstruction under viewpoint changes.}
(a) Renderings from increasing forward camera offsets and
(b) renderings at different camera heights within the same reconstructed
scene.
$d$ and $h$ denote the relative camera translation and height with
respect to the reference viewpoint.
}
    \textbf{}
    \label{fig:3dgs}
\end{figure}

\textbf{Baseline configurations.}
All five methods are evaluated using pretrained checkpoints without
additional fine-tuning.
ReCoSplat updates a scene through render-and-compare
\citep{cheng2026recosplat}, while
YoNoSplat supports both posed and unposed reconstruction
\citep{ye2026yonosplat}.
AnySplat and FreeSplatter-S jointly infer camera geometry and Gaussian
scene representations from unposed images
\citep{jiang2025anysplat,xu2025freesplatter}.
MVSplat uses posed multi-view cost volumes
\citep{chen2024mvsplat}; in our evaluation, it is supplied with
motion-integrated camera trajectories.

\textbf{Evaluation protocol.}
All methods are evaluated on rectified front-camera images from
200 \texttt{data\_test} clips.
Rendering quality is measured using 8-bit RGB images at
$224\times224$ resolution.
ReCoSplat, YoNoSplat, and AnySplat use 8--16 context views and are
evaluated on 2,495 held-out views in total.
FreeSplatter-S and MVSplat use two context views and are evaluated on
3,162 held-out views.
Metric values are first averaged over held-out views within each clip and
then equally averaged across clips.

Because the methods differ in context-view count, pose requirements,
native reconstruction interfaces, and held-out-view counts, the reported
results should be interpreted as system-level benchmark performance
rather than a controlled architecture-only comparison.

\textbf{Metric definitions.}
PSNR measures pixel-level reconstruction fidelity.
SSIM~\citep{ssim} and MS-SSIM~\citep{msssim} measure local and
multi-scale structural similarity, respectively.
LPIPS-VGG~\citep{zhang2018perceptual} measures learned perceptual distance,
while DISTS~\citep{ding2020dists} measures deep-feature structure and
texture differences.
FSIMc~\citep{fsim} combines phase, gradient, and color similarity,
whereas GMSD~\citep{gmsd} measures spatial variation in
gradient-magnitude similarity.
Together, these metrics cover complementary pixel-, structural-, and
perceptual-quality dimensions.

\textbf{Benchmark scope.}
The reconstruction task evaluates observation-grounded novel-view
rendering under viewpoint changes.
The reported metrics do not directly measure geometric reconstruction
accuracy, future scene dynamics, or closed-loop navigation safety.

\end{document}